\documentclass[letterpaper, 10 pt, conference]{ieeeconf}  

\IEEEoverridecommandlockouts                              

\usepackage{graphicx}
\usepackage{subcaption}
\usepackage{url}
\usepackage{amsmath}
\usepackage{amsfonts}
\usepackage{dsfont}

\usepackage{booktabs}
\usepackage{multicol, multirow}
\usepackage{rotating}
\usepackage{float}
\usepackage{hyperref}

\title{\LARGE \bf
Microsurgical Instrument Segmentation for Robot-Assisted Surgery
}

\author{Tae Kyeong Jeong, Garam Kim, and Juyoun Park$^*$
\\Korea Institute of Science and Technology, Seoul, South Korea%
\thanks{*corresponding author}
}

\begin{document}

\maketitle
\thispagestyle{empty}
\pagestyle{empty}

\begin{abstract}
Accurate segmentation of thin structures is critical for microsurgical scene understanding but remains challenging due to resolution loss, low contrast, and class imbalance. We propose \textit{Microsurgery Instrument Segmentation for Robotic Assistance(MISRA)}, a segmentation framework that augments RGB input with luminance channels, integrates skip attention to preserve elongated features, and employs an Iterative Feedback Module(IFM) for continuity restoration across multiple passes. In addition, we introduce a dedicated microsurgical dataset with fine-grained annotations of surgical instruments including thin objects, providing a benchmark for robust evaluation\footnote{Dataset available at \url{https://huggingface.co/datasets/KIST-HARILAB/MISAW-Seg}.}. Experiments demonstrate that MISRA achieves competitive performance, improving the mean class IoU by 5.37\% over competing methods, while delivering more stable predictions at instrument contacts and overlaps. These results position MISRA as a promising step toward reliable scene parsing for computer-assisted and robotic microsurgery.
\end{abstract}

\section{Introduction}
\label{intro}


Microsurgery(MS) is a surgical technique that manipulates blood vessels as small as 1-2\,mm in diameter and plays a critical role in lymphedema treatment and soft tissue reconstruction~\cite{badash2018supermicrosurgery, hong2018supermicrosurgery, hong2021supermicrosurgery}. While MS enables highly precise procedures with improved patient outcomes, it also demands exceptional dexterity and accuracy, as even minor errors can lead to complications~\cite{masia2014barcelona}. Thus, robot-assisted surgery has emerged to improve stability and accuracy in microsurgery~\cite{van2020first}. However, effective robotic assistance relies on accurate real-time segmentation of critical structures such as microvessels, needles, and wires~\cite{wang2023microsurgery}. Therefore, it is crucial to develop advanced segmentation methods tailored to the unique challenges of MS environments for enhanced robotic assistance in MS.


Despite its importance, MS scene segmentation faces two major challenges. First, there is a lack of publicly available datasets tailored for MS environments. Existing benchmarks such as EndoVis-2018~\cite{allan20192017} and CATARACTS~\cite{bouget2017vision} focus on endoscopic and ophthalmic surgeries and do not reflect the visual and magnification characteristics of MS. Moreover, no public dataset currently supports pixel-level segmentation for MS scenes, making it difficult to train or fairly evaluate models. We constructed the pixel-level segmentation dataset specifically designed for MS to address this gap, based on the MIcro-Surgical Anastomose Workflow recognition on training sessions(MISAW) dataset~\cite{huaulme2021micro}. Our dataset is essential not only for training robust segmentation but also for enabling fair and standardized benchmarking in future studies.


Secondly, another major challenge lies in accurately segmenting thin objects. Structures such as sutures and needles occupy only a few pixels in width, often exhibit low contrast, and are highly susceptible to detail loss during downsampling operations~\cite{adebar20143, stergiou2021refining}. Consequently, both CNN and transformer-based segmentation models struggle to preserve fine boundaries under these conditions, leading to reduced performance on MS images. Prior attempts such as Slim Scissors~\cite{han2022slim} addressed thin-object segmentation, but their reliance on interactive annotation makes them unsuitable for automated or real-time surgical scenarios.

\begin{figure}
    \centering    
    \includegraphics[width=0.75\linewidth]{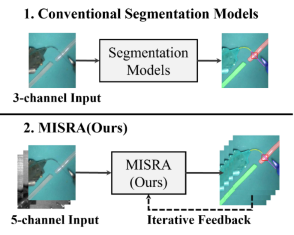}
    \caption{MISRA Framework}
    \label{framework}
\end{figure}

In this work, we propose a new segmentation framework, \textit{Microsurgery Instrument Segmentation for Robotic Assistance(MISRA)}, explicitly designed for thin-object recognition in MS scenes. Our approach incorporates three architectural innovations: 1) a \textbf{five-channel input representation} that augments RGB images with luminance channels to enhance the visibility of fine structures; 2) a \textbf{skip attention mechanism} that improves multi-scale feature fusion between encoder and decoder; and 3) an \textbf{iteration feedback module(IFM)} that injects high-resolution decoder features back into earlier encoder layers to reinforce thin-object representation. These design choices substantially improve segmentation accuracy for extremely thin surgical objects together with tailored loss functions.

Our contributions can be summarized as follows:
\begin{itemize}
    \item We develop and release the MS segmentation dataset with pixel-level annotations, enabling standardized evaluation of models on this domain.
    \item We propose \textit{MISRA}, a novel segmentation network that integrates five-channel inputs, skip attention, and an IFM to enhance thin object segmentation in MS images.
    \item We extensively evaluate our approach against other instance segmentation methods, including Mask2Former~\cite{cheng2022masked}, Mask R-CNN~\cite{he2017mask}, and U-Net~\cite{ronneberger2015u}, demonstrating consistent improvements particularly for thin objects.
    \item Our method improves the mean class IoU by 5.37\% over the strongest method, highlighting its effectiveness in MS environments.
\end{itemize}


\section{Related Work}
\label{related}
\subsection{Surgical Image Segmentation}


Segmentation of surgical scenes has been widely studied as a prerequisite for robot-assisted surgery, automated skill assessment, and post-operative analysis. Early approaches were dominated by CNNs, with architectures such as U-Net~\cite{ronneberger2015u}, Mask R-CNN~\cite{he2017mask}, and DeepLabV3+~\cite{chen2018encoder} forming the basis of many medical image segmentation pipelines. Recently, transformer-based models such as Mask2Former~\cite{cheng2022masked}, SegFormer~\cite{xie2021segformer} and Swin-UNet~\cite{cao2022swin} have demonstrated improved capacity for capturing global context, achieving strong performance in medical image analysis~\cite{yang2023tma}. 
Despite their success, these models are primarily optimized for macroscopic anatomical structures in endoscopic or laparoscopic settings. MS images differ significantly since they involve high magnification and extremely thin objects such as sutures and needles. Standard architectures often fail to capture such fine structures due to feature loss during downsampling, motivating the need for specialized approaches.

\subsection{Microsurgical Scene Datasets}



Research on microsurgical scenes remains underexplored compared to endoscopic segmentation. Rieke et al.~\cite{rieke2016real} introduced a retinal microsurgery dataset for instrument tracking, but it lacks segmentation annotations and does not cover vascular anastomosis. The MISAW dataset~\cite{huaulme2021micro} is more relevant, containing high-resolution images of robotic MS training sessions. However, it provides only workflow labels and kinematic data, without pixel-level segmentation masks. 
As highlighted by Rodrigues et al.~\cite{rodrigues2022surgical}, existing surgical tool datasets suffer from limited diversity, inconsistent annotation granularity, and a lack of standardization. In particular, most datasets emphasize workflow or instrument presence rather than dense pixel-level labels, making it difficult to conduct systematic segmentation research in microsurgical settings. This scarcity of standardized, annotated data highlights the need for a dedicated MS segmentation benchmark, which we address in this work.

\subsection{Thin Object Segmentation}




Accurately segmenting thin and elongated structures is a long-standing challenge in computer vision. Conventional CNN and transformer-based methods often lose boundary information due to repeated downsampling, which substantially limits their ability to capture fine details~\cite{long2015fully, marin2019efficient}. Several approaches have been proposed to mitigate this issue. Interactive methods such as Slim Scissors~\cite{han2022slim} and KnifeCut~\cite{lin2022knifecut} demonstrate that inpainting or cut-based feedback can effectively emphasize thin regions. However, their reliance on human input prevents scalability and real-time application in surgical settings. Boundary-aware networks such as NDNet~\cite{yang2020ndnet} attempt to improve thin object segmentation by aggregating multi-scale features to better preserve edge information. Other modules have explored complementary strategies, such as incorporating Sobel-based edge priors and feature denoising to enhance curvilinear boundaries~\cite{zhang2020befd}, or introducing boundary-aware loss functions that explicitly penalize misaligned contours~\cite{wang2022active}. While these methods contribute to improved boundary fidelity, they still struggle to robustly segment extremely thin, low-contrast structures such as sutures or needles in MS environments.
These limitations underscore the need for models explicitly designed to preserve fine structural information in MS scenes, which is the focus of our proposed method.

\section{Methodology}
\label{method}

\begin{figure}[t]
    \centering
    \includegraphics[width=0.95\linewidth]{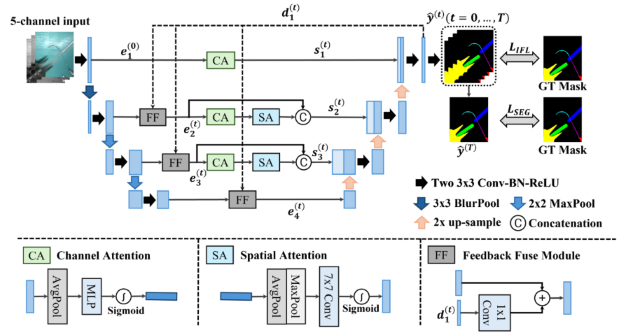}
    \caption{Overall architecture of MISRA}
    \label{misra}
\end{figure}

We propose \textit{MISRA}, a framework designed to address the challenges of thin object segmentation in MS scenes. MISRA combines a modified U-Net backbone, skip attention modules, an IFM, and tailored loss functions. The overall architecture is illustrated in Fig.~\ref{misra}.

\subsection{Notation}
Let $\Omega$ be the set of image input $I$ and $\mathcal{C}=\{1,\dots,C\}$ the class set.
At iteration index $t\in\{0,\dots,T\}$, the network outputs logits $z^{(t)}\in\mathbb{R}^{C\times H\times W}$ and class probabilities $p^{(t)}=\mathrm{softmax}(z^{(t)})$.
We denote the ground-truth label map by $y:\Omega\to\mathcal{C}$ and its one-hot encoding by $\mathbf{y}\in\{0,1\}^{C\times H\times W}$ with $\mathbf{y}_{c}(i) = [\,y(i) = c\,]$. 
We write $\mathrm{Dec}(\cdot)$ for the decoder, $\mathrm{Head}(\cdot)$ for the segmentation head, and $\mathrm{FF}_k(\cdot,\cdot)$, $\mathrm{CA}(\cdot)$, $\mathrm{SA}(\cdot)$ for Feedback-Fuse, Channel-Attention, and Spatial-Attention modules, respectively.
We denote the segmentation output at each iteration as $\hat{y}^{(t)}(i)=\arg\max_{c\in\mathcal{C}} p^{(t)}_{c}(i)$, with $\hat{y}^{(T)}$ being the final segmentation output.

\subsection{Input Preprocessing}
\label{preprocess}
We augment the RGB input with two luminance channels to enhance the visibility of thin, low-contrast structures. Given an RGB image $I\in[0,255]^{H\times W\times 3}$, we first convert it to grayscale $G\in[0,255]^{H\times W}$. Morphological erosion and dilation with a $4{\times}4$ structuring kernel $S$ produce the local minima and maxima maps ($m$, $M$). These are then min–max normalized to $[0,1]$, stacked as an additional two-channel tensor, and concatenated with the RGB channels to form the final 5-channel input. This process is summarized as follows:
\begin{equation}
\begin{aligned}
m &= G \ominus S,\ 
M = G \oplus S, \\
\tilde{m} &= \frac{m - \min(m)}{\max(m)-\min(m)+\varepsilon},\\
\tilde{M} &= \frac{M - \min(M)}{\max(M)-\min(M)+\varepsilon}, \\
X_{\mathrm{ext}} &= [\tilde{m},\,\tilde{M}] \in \mathbb{R}^{2\times H\times W}, \\
X_{\mathrm{in}} &= \mathrm{concat}\big(X_{\mathrm{RGB}},\,X_{\mathrm{ext}}\big)\in\mathbb{R}^{5\times H\times W}.
\end{aligned}
\label{luma}
\end{equation}
Here, $\ominus$ and $\oplus$ denote erosion and dilation, respectively. The extrema channels $\tilde{m}$ and $\tilde{M}$ provide local contrast cues that highlight thin edges, and concatenation with RGB yields the 5-channel input $X_{\text{in}}$ used throughout training and inference.

\subsection{Network Architecture}

\subsubsection{Encoder}
MISRA adopts a convolutional U-Net encoder with four stages. The 5-channel input (RGB + luminance channels) $X_{\text{in}}$ is mapped to 48 channels at the first stage, and the channel width doubles afterwards. Each stage comprises two $3{\times}3$ Conv–BN–ReLU blocks. Downsampling is performed by BlurPool~\cite{zhang2019making} at the first reduction to suppress aliasing that disproportionately harms thin boundaries and by $2{\times}2$ MaxPool at later reductions, which jointly improve robustness to high-frequency noise while preserving boundary integrity. 
Various architectures can be used for the convolutional encoder; for an ablation study, we substitute it with Pyramid Vision Transformer(PVT)~\cite{wang2021pyramid} and Swin Transformer~\cite{cao2022swin}.

\subsubsection{Skip Attention}
We apply attention at each skip connection to strengthen the fusion of features across different scales.
At the highest-resolution skip(s1), we apply Channel Attention(CA) only, since Spatial Attention(SA) at this level tends to amplify low-level noise. The intermediate skips(s2, s3) use CA + SA in a residual form to emphasize elongated surgical contours. We employ only a lightweight $3 \times3$ context block at the lowest resolution(e4)  to avoid oversmoothing fine features. 

\subsubsection{Decoder}
The decoder (Dec) follows a standard U-Net design with $2{\times}$ bilinear upsampling followed by $3{\times}3$ convolutional blocks. Features from the encoder are fused via skip connections.

\subsubsection{Iteration Feedback Module(IFM)}

Thin structures often vanish after a single forward pass due to downsampling and low contrast. To mitigate this, MISRA employs an IFM strategy inspired by high-resolution feedback networks ~\cite{hu2023high}. We denote by $e_k^{(t)}$ the feature of encoder stage $k\in\{1,2,3,4\}$ at iteration $t$, and by $s_j^{(t)}$ the attention-gated skip feature at resolution $j\in\{1,2,3\}$. $k\in\{1,2,3,4\}$ indexes encoder depth with $k{=}1$ the highest resolution and $k{=}4$ the lowest, and $j\in\{1,2,3\}$ indexes skip resolutions aligned with $e_1$, $e_2$, and $e_3$, respectively. At iteration $t$, the high-resolution decoder feature $d_1^{(t-1)}$ is fed back into encoder stages via Feedback Fuse(FF) modules, after which skip features are gated by attention, as follows:

\begin{equation}
\begin{aligned} \label{ifm}
e_k^{(t)} &= \mathrm{FF}_k\!\big(e_k^{(0)},\, d_1^{(t-1)}\big), \quad k\in\{2,3,4\},\\
s_1^{(t)} &= \mathrm{CA}\!\big(e_1^{(0)}\big),\\
s_2^{(t)} &= \mathrm{SA}\!\big(\mathrm{CA}(e_2^{(t)})\big) + e_2^{(t)},\\
s_3^{(t)} &= \mathrm{SA}\!\big(\mathrm{CA}(e_3^{(t)})\big) + e_3^{(t)}.
\end{aligned}
\end{equation}

The decoder then upsamples and fuses the gated skips with the context feature $e_4^{(t)}$ to generate a refined high-resolution feature $d_1^{(t)}$ and the corresponding logits $z^{(t)}$, as shown below:
\begin{equation}
\begin{aligned} \label{dec}
d_1^{(t)} &= \mathrm{Dec}\!\big(e_4^{(t)},\, s_3^{(t)},\, s_2^{(t)},\, s_1^{(t)}\big), \ 
z^{(t)} = \mathrm{Head}\!\left(d_1^{(t)}\right).
\end{aligned}
\end{equation}
The logits are then converted into class probabilities and discrete predictions, which are given by:
\begin{equation}
\begin{aligned} \label{out}
p^{(t)} &= \mathrm{softmax}\!\big(z^{(t)}\big),\  
\hat{y}^{(t)} = \underset{c}{\operatorname{argmax}}\; p_c^{(t)}.
\end{aligned}
\end{equation}
We set $T{=}3$ feedback iterations, yielding $T{+}1{=}4$ total passes including $t{=}0$. Intermediate predictions are supervised with iteration weights from~\cite{hu2023high}, defined as follows:
\begin{equation} \label{iter_weight}
\eta_t = w\cdot (t+1), \quad t=0,\dots,T.
\end{equation}
We set $w=0.1$, which yields $\eta=[0.1,\,0.2,\,0.3,\,0.4]$. 
The final segmentation is obtained from $\hat{y}^{(T)}$.

\subsection{Training Loss}
We train MISRA with two complementary objectives. The first supervises the final iteration output using class- and shape-aware segmentation losses. And the second regularizes all iterations in the IFM loop via weighted multi-iteration supervision with iteration weights $\eta_t$ from Eq.~\ref{iter_weight}. They encourage accurate thin object segmentation and stable refinement.

\subsubsection{Segmentation Loss}
We employ class weights $w_{\text{cls}}(c)$ computed by median frequency balancing(MFB)~\cite{eigen2015predicting} to mitigate class imbalance. We normalize these weights to have mean~1, referred to as \textit{normalized MFB(nMFB)}, which stabilizes training compared to raw MFB. We compare three settings in our ablation study: uniform weights (all ones), standard MFB, and nMFB.  

\noindent \textbf{Weighted Cross-Entropy Loss}\\ 
Eq.~\ref{ce_loss} defines the weighted Cross-Entropy(CE) loss, which enforces per-pixel correctness and is robust to class imbalance by using nMFB as class weights.
\begin{equation}
\begin{aligned}
L_{\mathrm{CE}}
= -\frac{1}{|\Omega|}\sum_{i\in\Omega} w_{\mathrm{cls}}\!\big(y(i)\big)\,\log p^{(T)}_{\,y(i)}(i))
\end{aligned}
\label{ce_loss}
\end{equation}

\noindent \textbf{Dice Loss} \\
We apply Dice loss~\cite{sorensen1948method} to complement CE loss, which directly optimizes region overlap and is particularly important when foreground objects occupy only a few pixels like needles and wires. In Eq.~\ref{dice_loss}, $p^{(T)}_{c}(i)$ and $\mathbf{y}_c(i)$ denote the predicted probability and one-hot indicator for class $c$ at pixel $i$, respectively.
\begin{equation}
\begin{aligned}
L_{\mathrm{Dice}}
= 1 - \frac{1}{C}\sum_{c\in\mathcal{C}}
\frac{2\sum_{i\in\Omega} p^{(T)}_{c}(i)\,\mathbf{y}_{c}(i)}
     {\sum_{i\in\Omega} p^{(T)}_{c}(i) + \sum_{i\in\Omega} \mathbf{y}_{c}(i) + \varepsilon}.
\end{aligned}
\label{dice_loss}
\end{equation}

\noindent \textbf{Focal Tversky Loss(FTL)} \\
Finally, we incorporate Focal Tversky Loss(FTL)~\cite{abraham2019novel} to emphasize thin classes. For class $c$, true positives, false positives, and false negatives are defined as $\mathrm{TP}_c=\sum_{i\in\Omega} p^{(T)}_{c}(i)\,\mathbf{y}_{c}(i)$, $\mathrm{FP}_c=\sum_{i\in\Omega} p^{(T)}_{c}(i)\,\big(1-\mathbf{y}_{c}(i)\big)$, and
$\mathrm{FN}_c=\sum_{i\in\Omega} \big(1-p^{(T)}_{c}(i)\big)\,\mathbf{y}_{c}(i)$. Eq.~\ref{ftl_loss} defines the Tversky index $TI_c$ and the FTL $L_{\mathrm{FTL}}$, where nMFB weights increase the contribution of thin classes. We follow ~\cite{abraham2019novel} and set $(\alpha,\beta,\gamma)=(0.3,\,0.7,\,4/3)$.
\begin{equation}
\begin{aligned}
\mathrm{TI}_c&=\frac{\mathrm{TP}_c}{\mathrm{TP}_c+\alpha\,\mathrm{FP}_c+\beta\,\mathrm{FN}_c+\varepsilon}, \\
L_{\mathrm{FTL}}&=\frac{1}{C}\sum_{c\in\mathcal{C}} w_{\mathrm{cls}}(c)\,\big(1-\mathrm{TI}_c\big)^{\gamma}.
\end{aligned}
\label{ftl_loss}
\end{equation}

\noindent \textbf{Combined Loss}\\
The overall segmentation loss combines these terms as shown in Eq.~\ref{seg_loss}. We set $\lambda_{\mathrm{CE}}{=}10$, $\lambda_{\mathrm{Dice}}{=}4$, and $\lambda_{\mathrm{FTL}}{=}0.3$.
\begin{equation}
\begin{aligned}
L_{\mathrm{SEG}}
= \lambda_{\mathrm{CE}}\, L_{\mathrm{CE}} \;+\; \lambda_{\mathrm{Dice}}\, L_{\mathrm{Dice}}
\;+\;\lambda_{\mathrm{FTL}}\, L_{\mathrm{FTL}}
\end{aligned}
\label{seg_loss}
\end{equation}

\subsubsection{Iteration Feedback Loss(IFL)}

We apply an auxiliary loss at every iteration to stabilize refinement across multiple iterations. Specifically, IFL computes class weighted CE loss  with $z^{(t)}$ and an IoU loss for each $t=0,\dots,T$~\cite{hu2023high}, balanced by iteration weights $\eta_t$ from Eq.~\ref{iter_weight}. The formulation is shown in Eq.~\ref{ifl_loss}. We set $\lambda_{\mathrm{CE}}=\lambda_{\mathrm{IoU}}{=}1$.
\begin{equation}
\begin{aligned}
L_{\mathrm{IFL}}
&= \sum_{t=0}^{T} \eta_t \Big[\, \lambda_{\mathrm{CE}}\, {L_{\mathrm{CE}}}^{(t)}
\;+\; \lambda_{\mathrm{IoU}}\, \big(1-\mathrm{mIoU}^{(t)}\big) \Big], \\
\mathrm{mIoU}^{(t)}
&= \frac{1}{C}\sum_{c\in\mathcal{C}}
\frac{\sum_{i\in\Omega} p^{(t)}_{c}(i)\,\mathbf{y}_{c}(i)}
     {\sum_{i\in\Omega}\big(p^{(t)}_{c}(i)+\mathbf{y}_{c}(i)-p^{(t)}_{c}(i)\,\mathbf{y}_{c}(i)\big)+\varepsilon}.
\end{aligned}
\label{ifl_loss}
\end{equation}

\subsubsection{Total Loss}
The final training loss is calculated as the sum of the segmentation loss and IFL, as shown below:
\begin{equation}
\begin{aligned}
L_{\mathrm{total}} \;=\; L_{\mathrm{SEG}} \;+\; L_{\mathrm{IFL}}.
\end{aligned} \label{total_loss}
\end{equation}

\section{Experiments}
\label{experiment}

\subsection{Experimental Setup}

\subsubsection{Dataset Construction}
\label{dataset}
As discussed in Section~\ref{intro}, one of the major obstacles in MS scene analysis is the lack of publicly available datasets specifically designed for segmentation. 
In response to this need, we constructed a new dataset based on the MISAW dataset~\cite{huaulme2021micro}. The MISAW dataset consists of synchronized kinematic and video data captured at 30 Hz using a master–slave robotic system developed by the University of Tokyo. The video data, acquired through a high-definition stereo microscope, yield left and right images at a resolution of 460×540 pixels. The dataset contains 27 training sessions conducted by 3 students and 3 expert surgeons, covering a wide range of skill levels. Each session includes both kinematic data and workflow annotations, such as surgical phases, procedural steps, and hand activities for each timestamp.

\begin{figure}[t]
    \centering    
    \includegraphics[width=0.95\linewidth]{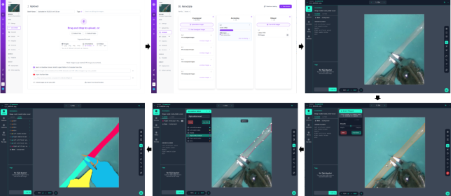}
    \caption{Annotation Flow for Dataset Construction}
    \label{roboflow}
    \vspace{-5pt}
\end{figure}

\begin{figure}[t]
\centering 
\begin{subfigure}{0.48\linewidth}
    \centering
    \includegraphics[width=\textwidth]{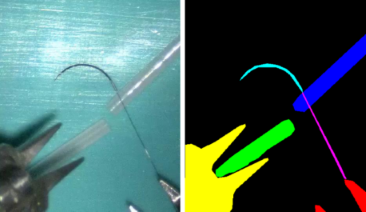}
    \label{sample1}
\end{subfigure}
\begin{subfigure}{0.48\linewidth}
    \centering
    \includegraphics[width=\textwidth]{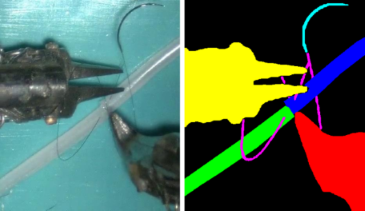}
    \label{sample2}
\end{subfigure}
\caption{Examples of Segmentation Label of MISAW Image}
\label{sample}
\end{figure}
    
We manually annotated MISAW images to create a segmentation dataset specifically tailored to the characteristics of MS. Annotations were performed using the Smart Polygon tool in Roboflow~\cite{roboflow}, which enabled accurate delineation of surgical instruments, needles, wires, and artificial vessels. 
As illustrated in Fig.~\ref{roboflow}, the annotation workflow ensured high-quality pixel-level masks for both thin and non-thin objects. The final dataset consists of 2,999 annotated images, divided into 2,433 training images and 566 test images, following the original MISAW protocol~\cite{huaulme2021micro} where data from one student and one expert are reserved for testing. 

The dataset includes six object classes: Left Artificial Vessel(LAV), Right Artificial Vessel(RAV), Left Needle Holder(LNH), Right Needle Holder(RNH), Needle(ND), and Wire(WR). Pixel-level statistics reveal that thin objects occupy only a small fraction of the image area: wires account for merely 3\% of all pixels and needles only 0.4\%. By contrast, larger structures such as artificial vessels and needle holders occupy approximately 15--30\% of the pixels. This extreme pixel-level sparsity, combined with the fine boundaries of thin structures, makes them the most challenging to segment and underscores the need for specialized models capable of fine-grained segmentation. Fig.~\ref{sample} shows examples of the image and annotations. The images on the left (first and third) are raw images from the MISAW dataset, while those on the right (second and fourth) are color masks representing the pixel-wise class annotations.

This dataset is one of the segmentation datasets tailored for MS scenes, providing a valuable resource for future research. Although constructed in a controlled training environment, it offers realistic high-magnification imagery of MS procedures and lays the groundwork for standardized evaluation of segmentation models targeting MS environments. The dataset introduced in this paper is publicly available at: \href{https://huggingface.co/datasets/KIST-HARILAB/MISAW-Seg}{huggingface.co/datasets/KIST-HARILAB/MISAW-Seg}.





\subsubsection{Evaluation Metrics}

We report a set of complementary metrics to capture both pixel-wise quality and instance-level behavior. The \textbf{mean class IoU(mcIoU)} is the arithmetic mean of per-class IoUs by averaging across classes. It prevents frequent categories from dominating the score and is therefore appropriate under the severe class imbalance and pixel sparsity of thin objects. The \textbf{ISI-IoU}~\cite{gonzalez2020isinet} follows prior surgical segmentation work and computes an instrument-sensitive IoU that reduces background bias and better reflects errors along tool boundaries. We also report \textbf{mean Dice(mDice)}, the Dice coefficient averaged over classes, which directly measures region overlap and is more stable when evaluating very small structures~\cite{azad2023loss}. 
\textbf{mAP@50} and \textbf{mAP@95} are mean average precision at IoU thresholds 0.50 and 0.95, standard in instance-segmentation benchmarks. 

\subsubsection{Implementation Details}

All models are implemented in PyTorch and trained on two NVIDIA RTX~A6000 GPUs (CUDA~11.3, PyTorch~1.12.1). We keep the native dataset resolution and do not use test-time augmentation. For \textit{MISRA}, we use AdamW with an initial learning rate of $5e{-4}$, weight decay of $1e{-3}$, 100 training epochs, and a batch size of 4. The 5-channel input (RGB + two luminance channels) follows Sec.~\ref{preprocess}, and the refinement loop uses $T{=}3$ feedback iterations with iteration weights $\eta=[0.1,0.2,0.3,0.4]$ as in Eq.~\ref{iter_weight}. Class weights are computed by nMFB on each training split. For other methods—U\textendash Net~\cite{ronneberger2015u}, Mask R\textendash CNN~\cite{he2017mask}, and Mask2Former~\cite{cheng2022masked}—we follow the setup of each original paper, while using the same nMFB class weights to handle severe class imbalance. Configurations of other methods are not forced to match MISRA so that each method is evaluated under its best practice setting.

\subsection{Comparison}
\label{comparison}

\begin{table}[h]
    \centering
    \caption{Comparison with other methods on MISAW dataset}
    \label{misaw_main}
    \scriptsize
    \setlength{\tabcolsep}{6pt}
    \begin{tabular}{l|ccccc}
        \toprule
        Method & mcIoU & ISI-IoU & mDice & mAP@50 & mAP@95 \\
        \midrule
        U-Net~\cite{ronneberger2015u} & 71.84 & 68.81 & 80.07 & 70.99 & 15.10 \\
        Mask R-CNN~\cite{he2017mask} & 68.25 & 69.92 & 75.82 & 74.39 & 4.2 \\
        Mask2Former~\cite{cheng2022masked} & \underline{77.13} & \underline{79.49} & \underline{83.24} & \underline{82.37} & \textbf{36.21} \\
        \midrule
        MISRA (ours) & \textbf{82.50} & \textbf{82.13} & \textbf{89.13} & \textbf{86.22} & \underline{31.20} \\
        \bottomrule
    \end{tabular}
\end{table}

As shown in Table~\ref{misaw_main}, MISRA surpasses other methods on the primary segmentation metrics. Against the strongest method Mask2Former~\cite{cheng2022masked}, MISRA improves mcIoU by +5.37\%, ISI-IoU by +2.64\%, mDice by +5.89\%, and mAP@50 by +3.85\%. mAP@95 is the only metric where another method leads: 36.21\% for Mask2Former and 31.20\% for MISRA. mAP@95 is largely dominated by large instruments with higher average overlaps, whereas thin objects rarely reach the 0.95 IoU threshold and thus contribute little to this metric. Accordingly, another method can appear slightly stronger on mAP@95, while MISRA remains superior on measures sensitive to thin objects. These results indicate that MISRA’s design—luminance-augmented inputs, skip attention, and an IFM—translates into stronger pixel-wise agreement and boundary adherence on the MS dataset.

\begin{table}[h]
    \centering
    \caption{IoU per class on MISAW dataset}
    \label{misaw_class}
    \scriptsize
    \setlength{\tabcolsep}{6pt}
    \begin{tabular}{l|cccccc}
        \toprule
        Method & LAV & LNH & ND & RAV & RNH & WR \\
        \midrule
        U-Net & 78.95 & \underline{93.05} & \underline{48.27} & 87.01 & 91.71 & 32.05 \\
        Mask R-CNN & 75.62 & 82.05 & 32.24 & 84.30 & 88.56 & 46.73 \\
        Mask2Former & \underline{82.94} & 92.57 & 44.90 & \textbf{90.90} & \textbf{94.65} & \underline{56.84} \\
        \midrule
        MISRA (ours) & \textbf{84.26} & \textbf{96.41} & \textbf{63.83} & \underline{90.37} & \underline{94.61} & \textbf{65.49} \\
        \bottomrule
    \end{tabular}
    \vspace{-5pt}
\end{table}

Class-wise results in Table~\ref{misaw_class} show that MISRA is particularly effective on thin classes(ND and WR). It achieves 63.83\% IoU on ND and 65.49\% on WR, improving over the strongest method by 15.56\% and 8.65\%, respectively. MISRA also attains the highest scores on LAV and LNH(84.26\% and 96.41\%) and remains competitive on RAV and RNH. These results indicate that MISRA reduces misses on thin structures and mitigates boundary fragmentation, yielding robust thin object segmentation without compromising performance on larger instruments.



\subsection{Ablation Study}

We analyze the contribution of each design choice on the MISAW dataset using four groups of experiments: encoder backbone, the number of feedback iterations, input channels and skip attention, and class weights with loss composition.

\begin{table}[h]
\centering
\caption{Ablation Study on Various Encoders}
\label{ablation-encoder}
\scriptsize
\setlength{\tabcolsep}{4pt}
\begin{tabular}{l|ccccc c}
\toprule
Encoder & mcIoU & ISI-IoU & mDice & mAP@50 & mAP@95 & param$\#$ \\
\midrule
Conv (default) & \underline{82.5} & \underline{82.1} & \underline{89.1} & \textbf{86.2} & 31.2 & \underline{6.10M} \\
PVT~\cite{wang2021pyramid}            & \textbf{83.0} & \textbf{83.0} & \textbf{89.4} & \underline{85.9} & \textbf{32.3} & 28.58M \\
Swin~\cite{cao2022swin}           & 81.6 & 81.4 & 88.3 & 84.8 & \underline{31.5} & \textbf{4.89M} \\
\bottomrule
\end{tabular}
\end{table}
\subsubsection{Encoder Backbone}
We evaluate the backbone choice by varying the encoder—Convolutional, PVT~\cite{wang2021pyramid}, and Swin~\cite{cao2022swin} (Table~\ref{ablation-encoder}). PVT attains the highest segmentation scores but with a substantially larger parameter budget, while Swin is the most compact yet lags across mcIoU, mDice, and mAP@50. The convolutional encoder offers a strong balance, delivering competitive accuracy with far fewer parameters. This trend suggests that, for thin surgical structures, precise local feature extraction and stable optimization are more decisive than heavier global modeling. We therefore adopt the convolutional backbone in MISRA as the accuracy–efficiency compromise.

\begin{table}[h]
\centering
\caption{Ablation Study on the Number of Feedback Iterations}
\label{ablation-iter}
\scriptsize
\setlength{\tabcolsep}{4pt}
\begin{tabular}{l|ccccc}
\toprule
Iteration $T$ & mcIoU & ISI-IoU & mDice & mAP@50 & mAP@95 \\
\midrule
0 (no feedback) & 80.8 & 79.9 & 87.8 & 84.3 & 28.7 \\
2               & 82.1 & 81.6 & 88.8 & 85.1 & 30.8 \\
3 (default)     & \textbf{82.5} & 82.1 & \textbf{89.1} & \textbf{86.2} & 31.2 \\
4               & \underline{82.4} & \textbf{82.8} & 89.0 & \underline{85.6} & \textbf{31.5} \\
5               & 82.4 & \underline{82.6} & \underline{89.1} & 85.5 & \underline{31.5} \\
\bottomrule
\end{tabular}
\end{table}
\begin{figure*}[h]
\begin{center}
\resizebox{\textwidth}{!}{
\renewcommand{\arraystretch}{0.2}
\begin{tabular}{c|ccccc}
\includegraphics[width=0.2\textwidth]{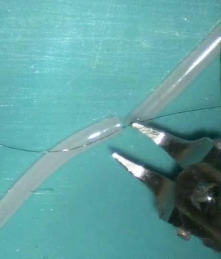}   &
\includegraphics[width=0.2\textwidth]{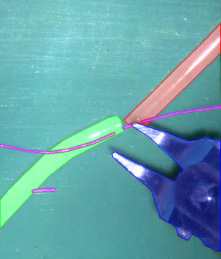}    &
\includegraphics[width=0.2\textwidth]{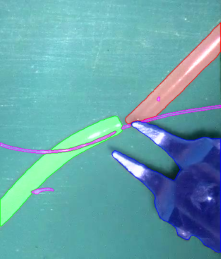} & \includegraphics[width=0.2\textwidth]{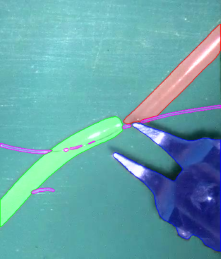}   &
\includegraphics[width=0.2\textwidth]{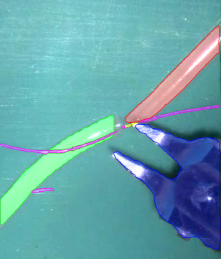} &
\includegraphics[width=0.2\textwidth]{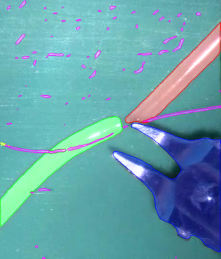}  \\

\includegraphics[width=0.2\textwidth]{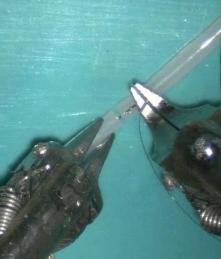}   &
\includegraphics[width=0.2\textwidth]{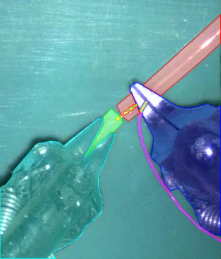}    &
\includegraphics[width=0.2\textwidth]{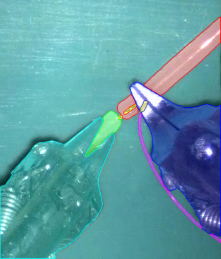} & \includegraphics[width=0.2\textwidth]{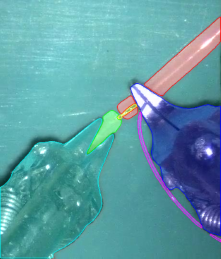}   &
\includegraphics[width=0.2\textwidth]{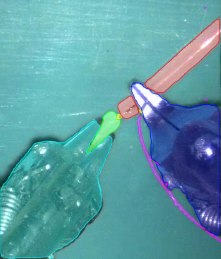} &
\includegraphics[width=0.2\textwidth]{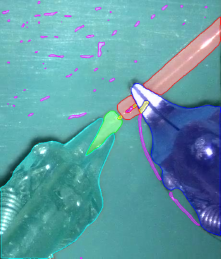}  \\

\includegraphics[width=0.2\textwidth]{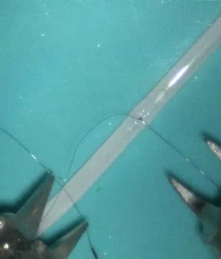}   &
\includegraphics[width=0.2\textwidth]{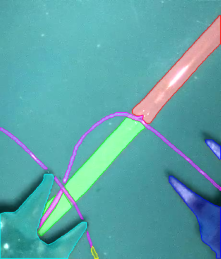}    &
\includegraphics[width=0.2\textwidth]{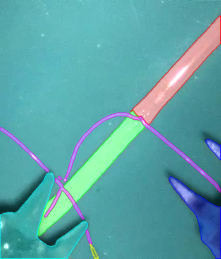} & \includegraphics[width=0.2\textwidth]{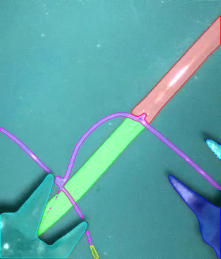}   &
\includegraphics[width=0.2\textwidth]{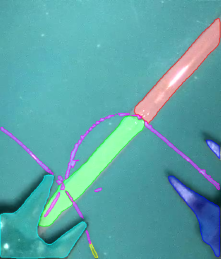} &
\includegraphics[width=0.2\textwidth]{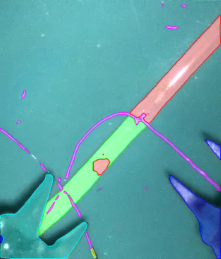}  \\

Original Image & GT & MISRA(Ours) & Mask2Former~\cite{cheng2022masked} & Mask R-CNN~\cite{he2017mask} & U-Net~\cite{ronneberger2015u} \\
\end{tabular}}
\renewcommand{\arraystretch}{1}
\end{center}
\caption{Visualization of Segmentation Results for Comparison on the MISAW Dataset}
\label{vis}
\end{figure*}

\subsubsection{The Number of Feedback Iterations}
We evaluate the role of the IFM by varying the number of iterations \(T\) (Table~\ref{ablation-iter}). Moving from \(T{=}0\) to \(T{=}3\) consistently improves all metrics, indicating that feature feedback helps recover missed thin segments and sharpen boundaries. Beyond three iterations, gains saturate: \(T{=}4\), \(5\) yield the highest mAP@95 but no clear advantage in mcIoU or mDice, likely due to diminishing returns and minor over-smoothing. We therefore set \(T{=}3\), balancing accuracy and computational cost.

\begin{table}[h]
\centering
\caption{Ablation Study on Input Channels and the Presence of Skip Attention}
\label{ablation-input-skip}
\scriptsize
\setlength{\tabcolsep}{4pt}
\begin{tabular}{ll|ccccc}
\toprule
Input & Skip attention & mcIoU & ISI-IoU & mDice & mAP@50 & mAP@95 \\
\midrule
RGB only     & Default                 & 81.8 & 81.6 & 88.6 & \underline{85.9} & \textbf{31.3} \\
RGB+Luma & None                  & 81.9 & \underline{81.8} & 88.7 & 85.6 & 30.6 \\
RGB+Luma & SA@all scales       & \underline{81.9} & 81.3 & \underline{88.7} & 85.3 & 29.3 \\
RGB+Luma & Default     & \textbf{82.5} & \textbf{82.1} & \textbf{89.1} & \textbf{86.2} & \underline{31.2} \\
\bottomrule
\end{tabular}
{\footnotesize Luma denotes the luminance channel, and SA denotes spatial attention.}
\end{table}
\subsubsection{Input Channels and Skip Attention}
We analyze luminance channels and skip attention placement in Table~\ref{ablation-input-skip}. Adding luminance channels to RGB improves mcIoU, mDice, and mAP@50, supporting the view that local contrast cues help delineate thin boundaries. Skip attention—channel attention at the highest resolution and channel attention + spatial attention at intermediate scales—further stabilizes fusion and yields the strongest overall performance. In contrast, applying spatial attention at all scales degrades accuracy, consistent with noise amplification at the finest resolution. These results motivate our default design: 5-channel inputs(RGB + Luminance channels) with skip attention.

\begin{table}[h]
\centering
\caption{Ablation Study on Class Weights and Loss Composition}
\label{ablation-weights-loss}
\scriptsize
\setlength{\tabcolsep}{4pt}
\begin{tabular}{ll|ccccc}
\toprule
Class Weights & Loss Comp. & mcIoU & ISI-IoU & mDice & mAP@50 & mAP@95 \\
\midrule
Uniform & Full               & \underline{82.4} & 81.6 & \underline{89.1} & 85.6 & \underline{31.3} \\
MFB     & Full               & 82.0 & 81.3 & 88.7 & \underline{86.1} & 30.6 \\
nMFB    & No FTL, IFL               & 82.1 & 81.7 & 88.8 & 84.8 & 31.2 \\
nMFB    & No FTL           & 81.7 & 81.2 & 88.5 & 85.3 & \textbf{31.9} \\
nMFB    & No IFL           & 82.3 & \underline{81.9} & 88.9 & 85.3 & 31.0 \\
nMFB    & Full (ours) & \textbf{82.5} & \textbf{82.1} & \textbf{89.1} & \textbf{86.2} & 31.2 \\
\bottomrule
\end{tabular}
{\footnotesize Full denotes the combination of CE, Dice, FTL, and IFL losses.}
\end{table}

\begin{figure*}[h]
\begin{center}
\resizebox{\textwidth}{!}{
\renewcommand{\arraystretch}{0.2}
\begin{tabular}{c|ccccc}
\includegraphics[width=0.2\textwidth]{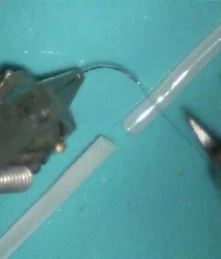}   &
\includegraphics[width=0.2\textwidth]{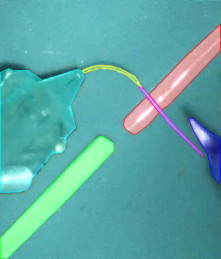}    &
\includegraphics[width=0.2\textwidth]{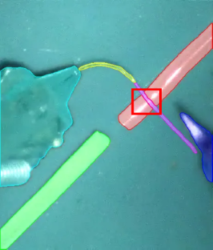} & \includegraphics[width=0.2\textwidth]{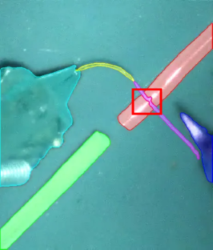}   &
\includegraphics[width=0.2\textwidth]{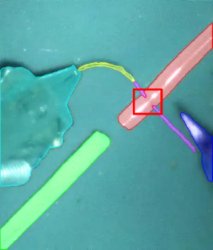} &
\includegraphics[width=0.2\textwidth]{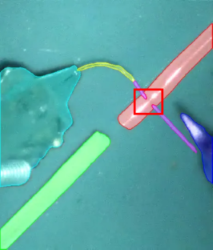}  \\

\includegraphics[width=0.2\textwidth]{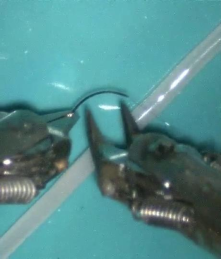}   &
\includegraphics[width=0.2\textwidth]{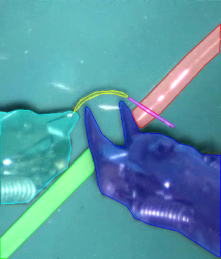}    &
\includegraphics[width=0.2\textwidth]{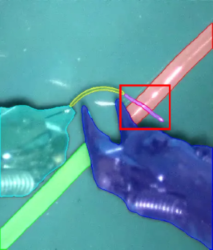} & \includegraphics[width=0.2\textwidth]{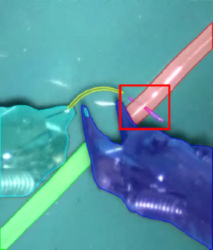}   &
\includegraphics[width=0.2\textwidth]{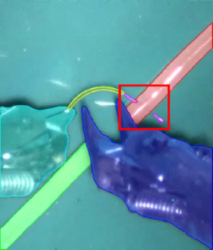} &
\includegraphics[width=0.2\textwidth]{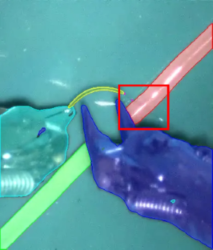}  \\

Original Image & GT & MISRA(Ours) & No Luminance Channel & No Skip Attention & No IFM \\
\end{tabular}}
\renewcommand{\arraystretch}{1}
\end{center}
\caption{Qualitative Ablation Study of MISRA Variants without Luminance Channels, Skip Attention, or IFM}
\label{vis_abl}
\vspace{-5pt}
\end{figure*}

\begin{figure*}[h]
\begin{center}
\resizebox{\textwidth}{!}{
\renewcommand{\arraystretch}{0.2}
\begin{tabular}{c|ccccc}
\includegraphics[width=0.2\textwidth]{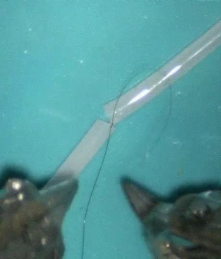}   &
\includegraphics[width=0.2\textwidth]{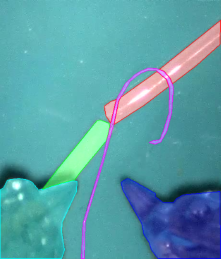}    &
\includegraphics[width=0.2\textwidth]{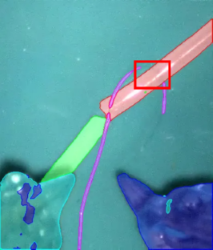} & \includegraphics[width=0.2\textwidth]{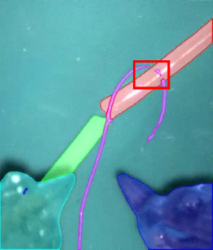}   &
\includegraphics[width=0.2\textwidth]{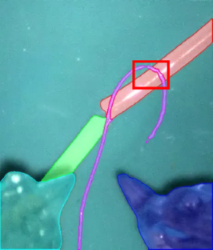} &
\includegraphics[width=0.2\textwidth]{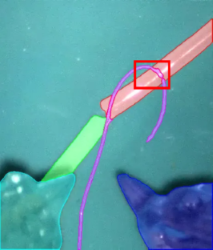}  \\

\includegraphics[width=0.2\textwidth]{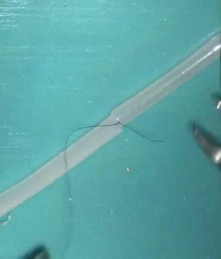}   &
\includegraphics[width=0.2\textwidth]{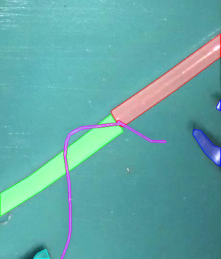}    &
\includegraphics[width=0.2\textwidth]{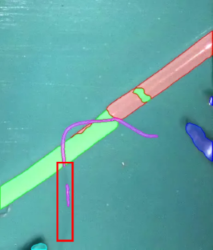} & \includegraphics[width=0.2\textwidth]{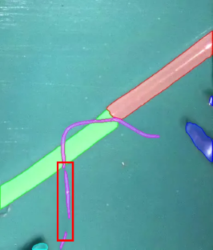}   &
\includegraphics[width=0.2\textwidth]{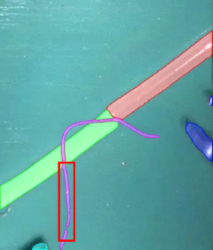} &
\includegraphics[width=0.2\textwidth]{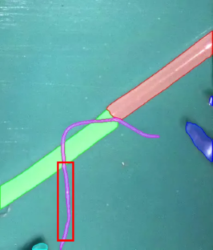}  \\

Original Image & GT & Itereration=0 & Itereration=1 & Itereration=2 & Itereration=3 \\
\end{tabular}}
\renewcommand{\arraystretch}{1}
\end{center}
\caption{Qualitative Visualization of MISRA across IFM Iterations ($T=0$–$3$)}
\label{vis_iter}
\end{figure*}

\subsubsection{Class Weighting and Loss Composition}
Table~\ref{ablation-weights-loss} evaluates two factors: 1) class weighting—Uniform to 1, raw MFB, and nMFB; and 2) loss composition—CE+Dice, CE+Dice+FTL, CE+Dice+IFL, and the full objective CE+Dice+FTL+IFL. Using nMFB yields the best overall balance compared with uniform and raw MFB, indicating more stable optimization under severe class imbalance. For the loss, CE+Dice forms a baseline; adding FTL improves overall segmentation accuracy, and IFL regularizes the multi-iteration refinement. Removing either term lowers at least one primary metric(e.g., mAP@50 drops to 85.3\% without IFL, while mcIoU/mDice fall to 81.7\%/88.5\%). Consequently, we adopt nMFB together with the full combination(CE+Dice+FTL+IFL) as the default setting.

\subsection{Visualization}
Fig.~\ref{vis} provides qualitative comparisons on MISAW. Across varied scenes, MISRA yields markedly more continuous masks for thin structures, especially wires. Thin structures remain connected, whereas other methods often show small breaks or missing segments. This property is attributed to the role of the IFM, which revisits high-resolution features across multiple iterations and repairs gaps that arise after a single forward pass.
Another recurring trend is robustness when thin objects are situated on top of other instruments. Predictions from other methods frequently fade or fragment in regions of contact, where wires or needles lie atop other instruments such as needle holders or artificial vessels. MISRA preserves thin object masks over heterogeneous backgrounds, suggesting that the skip attention provides the local context necessary to prevent absorption by neighboring classes.
Finally, MISRA produces slim and stable thin object masks. MISRA traces narrow, smoother boundaries that more closely match the ground-truth geometry than those produced by other models. We attribute this to the luminance-augmented input and a loss design that combines class-balanced weighting with shape-aware losses. These qualitative results are consistent with the class-wise improvements for thin objects shown in Table~\ref{misaw_class}.

Fig.~\ref{vis_abl} qualitatively assesses each component by comparing MISRA with variants that remove 1) luminance channels, 2) skip attention, and 3) IFM. Without luminance channels, masks are less stable: thin structures exhibit thickness jitter and ragged contours, indicating that luminance-augmented inputs help regularize fine boundaries. Without skip attention, thin objects over other instruments are absorbed by neighboring classes or locally erased, showing that skip attention preserves thin object signals on heterogeneous backgrounds. Without IFM, discontinuities increase and faint segments are missed more often. IFM closes gaps and stabilizes predictions where single-pass inference fails.

Fig.~\ref{vis_iter} visualizes the effect of the IFM in MISRA across different iterations. From $T{=}0$ to $T{=}3$, the highlighted regions show progressive restoration of connectivity, while boundaries become smoother without thickening.

\section{Conclusion}
\label{conclusion}
In this paper, we presented \textit{MISRA}, a segmentation framework tailored for the challenges of thin structures in MS scenes. Conventional segmentation architectures often fail to preserve features of thin objects through resolution reduction and struggle at contacts where thin structures lie on top of other instruments. MISRA addresses these failures by augmenting RGB input with luminance channels to enhance faint boundaries, applying skip attention to preserve elongated features, and refining predictions with an IFM that restores continuity across iterations.
In addition to the MISRA, we introduced a dedicated MS dataset with pixel-level annotations that explicitly include thin classes, providing a focused benchmark for  MS conditions. Comprehensive comparative experiments and ablations demonstrate that MISRA substantially improves thin-class accuracy, preserves connectivity at instrument contacts and crossings, and exhibits stable performance. Future works will focus on real-time deployment on robotic platforms and extending the approach to open-vocabulary surgical segmentation.










\bibliographystyle{IEEEtran}
\bibliography{cite}

\end{document}